\documentclass[sigconf]{acmart}
\usepackage{titlesec}
\titlespacing*{\section}{0pt}{\baselineskip}{\baselineskip}
\usepackage{multirow}
\usepackage{pifont}

\usepackage{graphicx}

\usepackage{booktabs} 
\usepackage{xspace}
\usepackage{makecell}
\usepackage{hyperref}
\usepackage{subcaption}
\hypersetup{
    colorlinks=true,
    linkcolor=blue,
    filecolor=magenta,      
    urlcolor=cyan
    }
\usepackage{enumitem}
\setlist{leftmargin=5mm}

\AtBeginDocument{%
  \providecommand\BibTeX{{%
    \normalfont B\kern-0.5em{\scshape i\kern-0.25em b}\kern-0.8em\TeX}}}

\begin{document}

\title{Towards Graph Foundation Models: A Transferability Perspective}


\author{Yuxiang Wang}
\email{yuxiawang@polyu.edu.hk}
\affiliation{%
  \institution{The Hong Kong Polytechnic University, HK SAR}
  \country{}}


\author{Wenqi Fan}
\email{wenqifan03@gmail.com}
\affiliation{
  \institution{The Hong Kong Polytechnic University, HK SAR}
  \country{}}

\author{Suhang Wang}
\email{szw494@psu.edu}
\affiliation{%
  \institution{The Pennsylvania State University, USA}
  \country{}}

\author{Yao Ma}
\email{may13@rpi.edu}
\affiliation{%
  \institution{Rensselaer Polytechnic Institute, USA}
  \country{}}


\begin{abstract}
    In recent years, Graph Foundation Models (GFMs) have gained significant attention for their potential to generalize across diverse graph domains and tasks. Some works focus on Domain-Specific GFMs, which are designed to address a variety of tasks within a specific domain, while others aim to create General-Purpose GFMs that extend the capabilities of domain-specific models to multiple domains. Regardless of the type, transferability is crucial for applying GFMs across different domains and tasks. However, achieving strong transferability is a major challenge due to the structural, feature, and distributional variations in graph data. To date, there has been no systematic research examining and analyzing GFMs from the perspective of transferability. To bridge the gap, we present the first comprehensive taxonomy that categorizes and analyzes existing GFMs through the lens of transferability, structuring GFMs around their application scope (domain-specific vs. general-purpose) and their approaches to knowledge acquisition and transfer. We provide a structured perspective on current progress and identify potential pathways for advancing GFM generalization across diverse graph datasets and tasks. We aims to shed light on the current landscape of GFMs and inspire future research directions in GFM development.
    
\end{abstract}







\maketitle

\section{Introduction}
\label{sec:1}
\vspace{-10pt}

\begin{figure}[]
  \centering
  \includegraphics[width=1\linewidth]{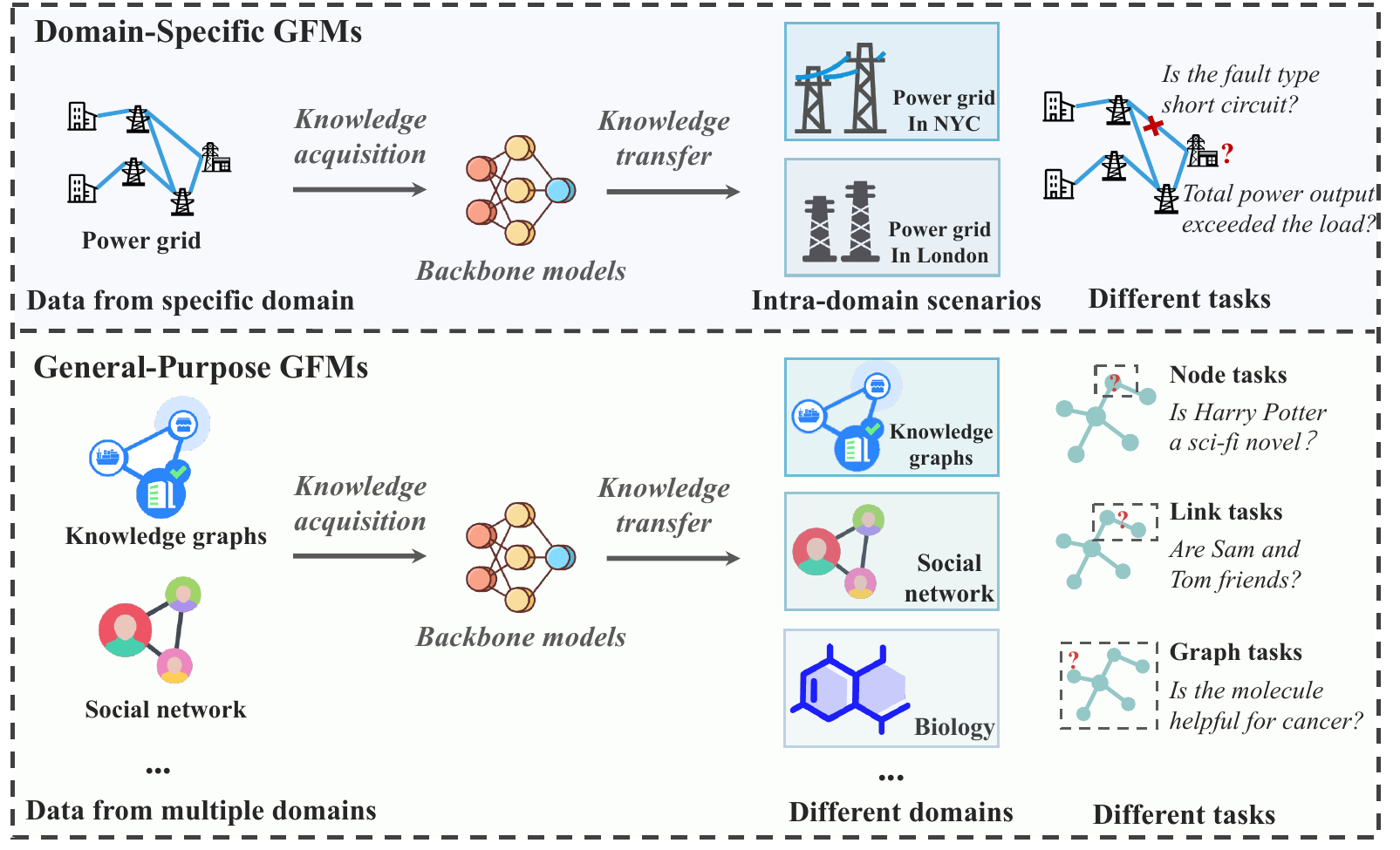}
  \caption{The application scope of GFMs. Domain-specific GFMs are designed to handle multiple graph tasks or intra-domain scenarios within a specific domain. General-purpose GFMs go beyond a single domain and extend the capabilities of domain-specific models to multiple domains.}
  \label{fig:GFM_bigpic}
\end{figure}

In recent years, large foundation models \cite{brown2020language, bommasani2021opportunities} have redefined artificial intelligence by shifting the focus from task-specific models to large-scale, pre-trained models that serve as versatile starting points for various applications. 
Through pre-training on vast amounts of data, large foundation models serve as a base model which can be adapted to a variety of downstream tasks \cite{bommasani2021opportunities, han2021pre}. These models have significantly advanced fields such as natural language processing \cite{bubeck2023sparks, touvron2023llama, achiam2023gpt}, computer vision \cite{radford2021learning, he2022masked, yuan2021florence}, and audio processing \cite{zhang2023video}.

Graphs, which represent entities as nodes and relationships as edges, are a fundamental data structure in many real-world applications \cite{wu2020comprehensive}. For instance, social networks model interactions between users \cite{graphsage}, molecular graphs encode chemical compounds for drug discovery \cite{gilmer2017neural}, and citation networks capture relationships between academic papers \cite{sen2008collective}. Graph-based tasks such as node classification (e.g., predicting interests of users in social networks), link prediction (e.g., recommending friendships in online platforms), and graph classification (e.g., identifying toxic molecules in drug discovery) are essential across these domains \cite{wu2022graph, gcn, hu2020open}. Inspired by the success of foundation models in other fields, researchers have begun exploring Graph Foundation Models (GFMs) as a way to bring similar benefits to graph-based data and tasks.

Recent advancements in GFMs have demonstrated their potential across a wide range of applications. 
Many efforts focus on \emph{domain-specific GFMs} \cite{huang2024foundation, wang2023transworldng, ULTRAQUERY}, which are designed to generalize within a single domain while adapting to various tasks and datasets.
As shown in Figure~\ref{fig:GFM_bigpic}, domain-specific GFMs extract essential graph-related knowledge from graphs within a particular domain (\emph{knowledge acquisition}) and transfer it across different datasets and tasks (\emph{knowledge transfer}).  
A successful example of a domain-specific GFM is TxGNN \cite{huang2024foundation}, designed for the medical domain. TxGNN predicts drug efficacy across various diseases, handling diverse drug types and generalizing to new medical conditions. By leveraging large-scale biomedical graphs, it offers a flexible, reusable model for drug discovery and personalized medicine. Similarly, in the power grid domain, researchers are actively developing GridFM~\cite{hamann2024foundation}, an ongoing initiative aimed at building a foundation model for power grids. GridFM aims to enable models trained for one region to adapt to grids in different cities and countries, helping to address challenges such as fault detection and peak load prediction.

While domain-specific GFMs focus on generalization within a specialized field, researchers are also exploring \emph{general-purpose GFMs} \cite{boog, xia2024anygraph,liu2023towards,mao2024graph}, which aim to extend beyond a single domain and transfer knowledge across different types of graphs.
As shown in Figure~\ref{fig:GFM_bigpic}, general-purpose GFMs acquire knowledge from multiple graph domains (\emph{knowledge acquisition}) and aims to transfer it across different domains and tasks (\emph{knowledge transfer}). Although building truly general-purpose GFMs remains an open challenge, recent advances have demonstrated promising steps in this direction \cite{xia2024opengraph, he2024unigraph, wang2024gft, qiu2020gcc, GCOPE}.
These improvements suggest that general-purpose GFMs are becoming more feasible, with early successes showing enhanced transferability across domains and tasks. If fully developed, these models could leverage shared graph structures across domains, potentially unlocking new opportunities in fields such as scientific and medical discovery.

A key factor in the success of GFMs—whether domain-specific or general-purpose—is their ability to transfer knowledge effectively. However, achieving strong transferability is challenging due to the structural, feature, and distributional differences across graphs \cite{maoposition}. GFMs rely on three essential types of transferability. 1) \emph{Intra-domain transferability} allows models to adapt to new datasets within the same domain, even when data distributions vary; 2) \emph{Cross-domain transferability} enables general-purpose GFMs to apply knowledge from one domain to another; and 3) \emph{Cross-task transferability} allows models to adapt to different learning objectives, such as shifting from node classification to link prediction. For domain-specific GFMs, intra-domain and cross-task transferability are key to ensuring adaptability across datasets and tasks within the same field. On the other hand, general-purpose GFMs must also excel at cross-domain transferability, as they need to generalize across fundamentally different graph domains.

In this paper, we provide a taxonomy of existing GFMs based on their application scope (domain-specific vs. general-purpose), and their focus on different types of transferability. The life cycle of GFMs consists of three key components that collectively enable good transferability: 1) \emph{The Backbone Model}, which serves as the foundation for learning and representing graph data. A well-designed backbone must be capable of capturing rich and transferable knowledge to support adaptation across tasks and domains; 2) \emph{Knowledge Acquisition}, which prepares the backbone by extracting meaningful and generalizable patterns from graph data using various learning techniques; and 3) \emph{Knowledge Transfer}, which determines how well the acquired knowledge is applied to new tasks or domains. Effective transfer strategies allow GFMs to adapt flexibly, overcoming structural and distributional differences in graph data. We review existing GFMs through the lens of these components, summarizing their architectural designs, learning strategies, and transfer mechanisms.

\noindent{\textbf{Comparison with Existing Surveys for GFMs.}}
Several recent surveys have explored aspects of GFMs, but they differ in scope from our work. \cite{wu2025graph} focuses specifically on GFMs for recommendation systems. \cite{liu2023towards} is a comprehensive survey on GFMs, which categorizes GFMs primarily based on their backbone architectures, such as GNN-based and LLM-based models.  \cite{maoposition} is a position paper that introduces the concept of a unified graph vocabulary to improve cross-domain transferability, emphasizing the importance of structured representations in GFMs rather than a complete survey of existing models. Lastly, \cite{zhao2024survey} focuses on self-supervised learning for GFMs, particularly from a knowledge-based learning perspective. In contrast, our survey takes a transferability-centric perspective, systematically analyzing how existing GFMs achieve transferability. By structuring GFMs around their application scope (domain-specific vs. general-purpose) and their approaches to knowledge acquisition and transfer, we provide a structured perspective on current progress and identify potential pathways for advancing GFM generalization across diverse graph datasets and tasks.

\section{A Taxonomy of GFMs}
\label{sec:2.1}
\vspace{-10pt}
GFM is a powerful paradigm for learning transferable representations of graph-structured data, but not all GFMs share the same objectives. Some GFMs \cite{ULTRAQUERY, huang2024foundation, wang2023transworldng} focus on achieving strong performance within a specific domain, leveraging domain-specific properties to optimize task accuracy. In contrast, others \cite{xia2024anygraph, xia2024opengraph, wang2024gft} aim to provide a broad, generalizable framework that can be applied to diverse graph domains with minimal adaptation. To provide a systematic understanding of GFMs, we introduce a taxonomy that categorizes models based on two dimensions: (1) application scope, which differentiates between domain-specific and general-purpose GFMs, and (2) transferability capabilities, which define how well a model generalizes across datasets, domains, and tasks. 
In this way, this taxonomy clarifies the strengths and limitations of different GFMs, highlighting the types of transferability required for effective generalization.

\vspace{-5pt}
\subsection{Domain-Specific vs. General-Purpose GFMs}
Based on the application scope, GFMs can primarily be divided into domain-specific GFMs and general-purpose GFMs, as shown in Figure \ref{fig:GFM_bigpic}. \emph{Domain-specific GFMs} are designed to handle multiple graph tasks or intra-domain scenarios within a specific domain. For instance, GridFM \cite{hamann2024foundation} is a power grid GFM that can be applied to power systems in different countries and cities. Additionally, it can also tackle various tasks within the grid system, such as fault detection and peak load detection. In contrast, \emph{general-purpose GFMs} go beyond a single domain and extend the capabilities of domain-specific models to multiple domains. For example, AnyGraph \cite{xia2024anygraph} is a versatile GFM designed to handle diverse graph data and tasks efficiently, showcasing strong generalization capabilities across different domains (e.g. e-commerce and academic domains). General-purpose GFMs aim to transfer knowledge across domains with minimal adaptation.

\vspace{-5pt}
\subsection{Transferability: Core Requirement for GFMs}
\label{sec:2.2}
To achieve their intended objectives, GFMs must exhibit different types of transferability. 
We identify three key forms of transferability that play a critical role in GFM performance: 1) {\bf Intra-Domain Transferability}: the ability to generalize within the same domain, even when trained and tested on datasets with different distributions; 2) {\bf Cross-Domain Transferability}: the ability to apply learned representations from one domain to another. This is the defining challenge for general-purpose GFMs, as they aim to generalize across diverse graph structures while maintaining performance; and 3) {\bf Cross-Task transferability}: The ability to adapt a pretrained model to different tasks, such as transitioning from node classification to link prediction. Both domain-specific and general-purpose GFMs benefit from task-level transfer, as it enhances model versatility and reusability.

In general, domain-specific GFMs focus primarily on two aspects of transferability: intra-domain transferability and cross-task transferability. Intra-domain transferability ensures that the model performs well across different datasets within the same domain, while cross-task transferability allows the model to generalize across various tasks. For example, a power grid GFM \cite{hamann2024foundation} needs intra-domain transferability to adapt to diverse power grid systems across different countries and cities, each with its own data distribution. Additionally, it requires cross-task transferability to handle tasks such as fault detection and peak load detection effectively. 
On the other hand, general-purpose GFMs are designed to go beyond a single domain and achieve cross-domain transferability. This means that they can operate across different types of graph data and transfer learned knowledge from one domain to another, making them a versatile solution for a broader range of applications.

\begin{table}[h!]
\caption{Taxonomy of Graph Foundation Models}
\label{tab:Boundaries of GFMs}
\centering
\scalebox{0.6}{%
\begin{tabular}{c l c c c}
\toprule

 \makecell{\textbf{Application} \\ \textbf{scope}} & \textbf{Paper} & \makecell{\textbf{Intra-domain} \\ \textbf{transferability}} & \makecell{\textbf{Cross-task} \\ \textbf{transferability}} & \makecell{\textbf{Cross-domain} \\ \textbf{transferability}}  \\
\midrule
\multirow{13}{*}{\bf Domain-Specific} & Prodigy \cite{huang2024prodigy} & - & \ding{51} & \\
&InstructGLM \cite{InstructGLM} &- & \ding{51}& \\

&GraphGPT \cite{tang2024graphgpt} &- & \ding{51}& \\
&ULTRA \cite{galkin2023towards} & Knowledge graph& \ding{51}& \\

&ULTRAQUERY \cite{ULTRAQUERY} & Knowledge graph& \ding{51}& \\
&MiniMol \cite{klaser2024minimol} & Biology domain & \ding{51}& \\

&COATI \cite{kaufman2024coati} &Biology domain & \ding{51}& \\
&JMP \cite{jmp} &Biology domain &\ding{51} & \\

&Triplet-GMPNN \cite{ibarz2022generalist} & Algorithmic reasoning &\ding{51} & \\
&TxGNN \cite{huang2024foundation} &Medicine domain &\ding{51} & \\

&TFM \cite{wang2023building} & Transportation domain &\ding{51} & \\
&TransWorldNG \cite{wang2023transworldng} &Transportation domain &\ding{51} & \\

&GridFM \cite{hamann2024foundation} &Power grid domain &\ding{51} & \\

\midrule

\multirow{14}{*}{\bf General-Purpose} & BooG \cite{boog} &- & \ding{51} & \ding{51} \\
&AnyGraph \cite{xia2024anygraph} &- &\ding{51} &\ding{51} \\

&GraphFM \cite{lachi2024graphfm} &- & &\ding{51} \\
&GraphAny \cite{zhao2024graphany} & - & &\ding{51} \\

&OpenGraph \cite{xia2024opengraph} & - & \ding{51}&\ding{51} \\
&Unigraph \cite{he2024unigraph} &  -&\ding{51} &\ding{51} \\

&OFA \cite{liu2023one} & - &\ding{51} &\ding{51} \\

&ProG \cite{sun2023all} & - &\ding{51} &\ding{51} \\
&ZeroG \cite{li2024zerog} &  -& &\ding{51} \\

&GCOPE \cite{GCOPE} & - & &\ding{51} \\
&MDGPT \cite{MDGPT} & - &\ding{51} &\ding{51} \\

&GCC \cite{qiu2020gcc} &  -&\ding{51} &\ding{51} \\   
&LLaGA \cite{chen2024llaga} & - & &\ding{51} \\

&GFT \cite{wang2024gft} & - & \ding{51}&\ding{51} \\

\bottomrule
\end{tabular}%
}

\end{table}
\vspace{-1pt}
\subsection{Unified Taxonomy of GFMs}
To provide a clearer understanding of existing GFMs, Table~\ref{tab:Boundaries of GFMs} presents a taxonomy that categorizes models based on their application scope (domain-specific vs. general-purpose) and their transferability capabilities. Next, we briefly describe the existing GFMs.

For Domain-Specific GFMs, many have shown impressive performance in real-world applications, demonstrating strong intra-domain and cross-task transferability. For instance, ULTRA \cite{galkin2023towards} and ULTRAQUERY \cite{ULTRAQUERY} focus on the knowledge graph domain, excelling in link prediction and logical query answering. In the biology domain, MiniMol \cite{klaser2024minimol}, COATI \cite{kaufman2024coati}, and JMP \cite{jmp} have made significant strides by improving predictive models for molecular properties and interatomic potentials, which has advanced drug discovery and material science. Similarly, TxGNN\cite{huang2024foundation} has advanced drug repurposing by generating actionable predictions for diseases with limited treatment options. Beyond biology, Triplet-GMPNN \cite{ibarz2022generalist} improves performance across various algorithmic tasks. In transportation, TFM \cite{wang2023building} and TransWorldNG \cite{wang2023transworldng} enhance simulations for better traffic management. Models like Prodigy\cite{huang2024prodigy}, InstructGLM\cite{InstructGLM}, and GraphGPT~\cite{tang2024graphgpt} are not designed for any specific domain but rather operate at the methodology level, providing general techniques that can be applied across different tasks.

General-purpose GFMs remain an area of ongoing research and have yet to be directly applicable in real-world scenarios. Current efforts focus on exploring more effective strategies for cross-domain transfer. For example, BooG \cite{boog}, AnyGraph \cite{xia2024anygraph}, Unigraph \cite{he2024unigraph}, OFA \cite{liu2023one}, and ProG \cite{sun2023all} unify the structural and feature characteristics from diverse graph data, enabling them to handle tasks across various domains, from social network analysis to chemical molecule prediction. OpenGraph \cite{xia2024opengraph}, pre-trained on large-scale generated graphs, demonstrates strong zero-shot performance across domains such as social and academic networks. GFT \cite{wang2024gft} uses a transferable tree vocabulary, facilitating the effective handling of tasks in domains like social networks and molecular graphs. While GraphFM \cite{lachi2024graphfm}, GraphAny \cite{zhao2024graphany}, ZeroG \cite{li2024zerog}, and GCOPE \cite{GCOPE} have not achieved cross-task transferability, they still perform well in node classification across different domains. Additionally, most existing general-purpose GFMs do not explicitly focus on intra-domain transferability, especially when handling distribution shifts within the same domain. While they emphasize cross-domain adaptability, their generalization within a single domain under varying conditions remains underexplored.

\section{Overall Framework of GFMs}
\label{sec:2.3}
\vspace{-10pt}

To achieve transferability, GFMs must first learn and retain transferable knowledge (e.g. transferable graph patterns) from graph data. Building on this foundation, they can then apply relevant knowledge to new tasks or domains as needed. In particular, as illustrated in Figure~\ref{fig:GFM_bigpic}, a general framework of GFMs consists of three main components: 1) The \emph{Backbone Model} provides the fundamental architecture for processing graph data and learning representations; 2) \emph{Knowledge Acquisition} equips GFMs with transferable graph patterns by leveraging various learning techniques; and 3) \emph{Knowledge Transfer} ensures that learned knowledge can be effectively applied to different tasks or domains.

At the core of GFMs is the \emph{Backbone Model}, which serves as the foundation for learning graph representations. A strong backbone must be capable of capturing key graph properties, including structural relationships, node features, and semantic information. Without a well-designed backbone, a GFM may fail to learn the necessary representations for effective transfer. 
Once the backbone is established, \emph{Knowledge Acquisition} enables GFMs to extract meaningful patterns from graph data. This step involves learning from diverse datasets and applying techniques such as supervised learning \cite{nasteski2017overview}, contrastive learning \cite{you2020graphcl, dgi}, and generative modeling \cite{hou2022graphmae, hou2023graphmae2}. Through these methods, GFMs build a knowledge base that helps them adapt to new tasks. As shown in Figure~\ref{fig:GFM_bigpic}, domain-specific GFMs acquire knowledge from a single domain, such as power grids, while general-purpose GFMs integrate knowledge from multiple domains, including social networks, biology, and knowledge graphs. After acquiring knowledge, \emph{Knowledge Transfer} determines how well a GFM applies what it has learned to new tasks or domains. Since different tasks emphasize distinct aspects of graphs, and data distributions can vary significantly across domains, knowledge transfer focuses on overcoming these challenges and enabling rapid and effective application of the learned knowledge to downstream tasks.

In the following sections, we will review existing GFMs through the lens of these three components, summarizing their design choices and strategies for improving transferability.

\textbf{}
\section{Backbone Model}
\label{sec:3}
\vspace{-10pt}

Backbone models form the structural foundation of GFMs and play a critical role in their ability to learn transferable knowledge from graph data. Given the diversity of graph structures and features, GFMs are expected to learn useful and transferable graph patterns. This requires the backbone model to have sufficient \emph{expressiveness} to embed complex features and accurately represent diverse topologies \cite{zhang2023expressive}. Without expressiveness, even if graph data contains rich patterns, the GFM will lack the ability to capture and utilize them effectively. Moreover, since GFMs are expected to be applicable across multiple tasks or domains, they cannot undergo significant changes when learning or adapting to unseen structures and features, as this could degrade their performance in other tasks or domains. Therefore, the backbone model needs to demonstrate \emph{flexibility}, an intrinsic ability to handle different graph structures, tasks, and features without requiring substantial architectural modifications. Additionally, GFMs need to build large models by consuming vast amounts of data. This requires the backbone model to possess a degree of \emph{scalability}, allowing GFMs to handle larger datasets and potentially exhibit improved performance and emergent abilities as they scale~\cite{xia2024anygraph, lachi2024graphfm, ma2022graph}. In the following subsection, we will discuss the expressiveness, flexibility, and scalability of commonly used graph learning architectures. We then explore how current GFMs utilize these backbone models.

\vspace{-5pt}
\subsection{Common Graph Learning Architectures}
\label{sec:3.2}
The backbone models of current GFMs generally follow common graph learning architectures, which have evolved to handle increasing complexities and a wide variety of graph tasks. The progression begin with shallow embedding models, followed by GNN-based models, then transformer-based models, and more recently, LLM-based models. Shallow embedding models \cite{deepwalk, node2vec, tang2015line} learn low-dimensional node representations based on local graph structures. However, their limited expressiveness and flexibility have made them less common in GFMs. GNN-based models \cite{gcn, graphsage} introduced the message-passing paradigm, enabling nodes to aggregate information from neighbors, offering improved expressiveness. Transformer-based models \cite{graphbert, graphormer} leverage the self-attention mechanism to capture both local and global dependencies. Most recently, LLM-based models \cite{InstructGLM, tang2024graphgpt, chen2024llaga} use pre-trained language models to process graph data by converting graphs into token sequences, offering even greater power in handling graph data. This evolution reflects the increasing demand for models that are expressive, flexible, and scalable. In this section, we primarily focus on the most representative models within the GNN-based, transformer-based, and LLM-based architectures and discuss their capabilities, such as GCN \cite{gcn} from the GNN-based architecture. Although more advanced models \cite{zhou2022mentorgnn, Geom-GCN, chaudhary2024gnndld} built on these architectures show promise, they are less commonly adopted in GFMs. This is likely due to their increased complexity and task-specific design. We will leave a detailed discussion of these models for future work.

\vspace{-5pt}
\subsubsection{GNN-based Models.}
\label{sec:3.2.2}
GNN-based models \cite{gcn, gat, gin} are the most popular deep learning architectures for graphs \cite{zhang2020deep}. Most representative GNNs adopt a message-passing paradigm, where nodes exchange messages with their neighbors to update their representations \cite{zhang2023graph}. They aggregate multi-hop information and make use of node features, allowing them to effectively express various graph structures and features. Nevertheless, the locality bias restricts their capacity to transfer knowledge for tasks demanding a deeper comprehension of global structures, thereby resulting in expressiveness bounds \cite{morris2019weisfeiler, gin}. They have limited flexibility since their reliance on local aggregation limits their fast and effective knowledge transfer across different tasks and domains \cite{sun2311graph}. Besides, they face challenges with scalability due to their iterative message-passing nature. Techniques like sampling \cite{graphsage, chen2017stochastic, chen2018fast, zeng2019graphsaint} help, but the locality bias limits how well they transfer knowledge to large, complex graphs.

\vspace{-5pt}
\subsubsection{Transformer-based Models.}
\label{sec:3.2.3}
Transformer-based models \cite{graphormer, graphbert, gtn, TokenGT, SAN} adapt standard Transformers \cite{transformer} for graph data by treating graph nodes as tokens and using attention mechanisms to capture both local and global dependencies between nodes. TokenGT \cite{TokenGT} and SAN \cite{SAN} further explore how to tokenize additional graph structural information. This approach makes transformer-based models highly expressive, enabling them to effectively transfer complex relational information to downstream tasks. They are also more flexible than GNN-based models, capable of handling diverse graph types. However, their reliance on attention mechanisms may hinder performance on simpler tasks where local aggregation is sufficient. Additionally, their scalability is limited due to the high computational cost of global attention, which reduces their transferability to larger graphs \cite{zhang2023graph, lim2024benchmarking}.

\subsubsection{LLM-based Models.}
\label{sec:3.2.4}
LLM-based models \cite{jin2024large, chen2024exploring, guo2023gpt4graph, zhu2025llm, zhao2023graphtext, wang2023graph, wang2024instructgraph, qin2023disentangled} use large language models to process graph data by converting graphs into natural language-like sequences. This conversion typically involves representing graph structures and node relationships in a textual format, where nodes are treated as entities and edges as relationships between them \cite{liu2023towards}. By using the powerful language understanding capabilities of LLMs, these models capture complex semantic relationships in graphs and apply natural language reasoning to graph-based tasks.
They generally have good expressiveness but their focus on sequential data limits their ability to efficiently represent and transfer graph structures. Besides, they are highly flexible, as they can be applied to various graph tasks once the graph is converted into token sequences. However, their scalability is hindered by the inefficiency of transforming graph data into sequences, making them less suited for very large graphs.

\subsection{Backbone Models of Existing GFMs}
\label{sec:3.4}

In this section, we discuss the backbone of existing GFMs. 
\vspace{-5pt}
\subsubsection{GNN-based models.} GNN-based models are the predominant backbone for current GFMs, valued for their ability to handle relational data and facilitate message passing between nodes, which is crucial for capturing structural context in graphs. Models like OFA \cite{liu2023one}, Prodigy \cite{huang2024prodigy}, ProG \cite{sun2023all}, and GCOPE \cite{GCOPE} all leverages expressiveness of GNNs to capture node relationships and contextual information within graphs. GCC \cite{qiu2020gcc}, MiniMol \cite{klaser2024minimol}, and JMP \cite{jmp}, on the other hand, focus on pre-training GNNs to develop transferable structural representations. 
Some GFMs enhance GNN flexibility and expressiveness by adding specialized structures to handle diverse graph data. For example, AnyGraph \cite{xia2024anygraph} addresses cross-domain heterogeneity by leveraging GNNs in combination with a Mixture-of-Experts (MoE) architecture, where each expert model specializes in handling graphs with distinct characteristics. An automated routing algorithm assigns each graph to the most suitable expert, enabling AnyGraph to effectively learn transferable knowledge from diverse graph structures and adapt to new domains and distribution shifts. GraphAny \cite{zhao2024graphany} generalizes node classification across graphs with different feature and label spaces. It models inference as an analytical solution to a LinearGNN, allowing flexible adaptation to new graphs. An inductive attention module combines predictions from multiple LinearGNNs, weighting each based on entropy-normalized distances, which enhances expressiveness and transferability across diverse graph domains.

\vspace{-5pt}
\subsubsection{Transformer-based models.} As graph data grows in complexity, transformer-based models are becoming popular due to their ability to capture intricate relationships via self-attention. For example, GraphFM \cite{lachi2024graphfm} introduces a perceiver-style transformer, which compresses domain-specific graph features into a shared latent space. By using learned latent tokens, GraphFM captures complex structural patterns across graphs, allowing it to perform well across multiple datasets with minimal fine-tuning. OpenGraph \cite{xia2024opengraph}, aiming for zero-shot transfer across datasets, employs a unified graph tokenizer that converts any graph into universal tokens, processed by a scalable transformer to capture node-wise dependencies. 

\vspace{-5pt}
\subsubsection{LLM-based models.} 
LLM-based models are increasingly recognized as powerful tools in the design of GFMs. Leveraging the capabilities of LLMs, they have shown remarkable performance across various graph tasks. For instance, InstructGLM \cite{InstructGLM} and LLaGA \cite{chen2024llaga} have set new benchmarks, surpassing GNN-based and transformer-based models on popular datasets like Cora and PubMed. GraphGPT \cite{tang2024graphgpt} takes this a step further by incorporating a text-graph grounding component, which aligns textual descriptions with graph topologies. This unique feature enables the model to effectively process and interpret intricate graph structures.

\vspace{-5pt}
\subsubsection{Hybrid models.} Hybrid models combine GNNs with transformers or pre-trained language models (PLMs) to capture both structural and semantic information. In these models, GNNs handle structural dependencies within graphs, while transformers or PLMs process sequential or semantic data. For instance, COATI \cite{kaufman2024coati} uses GNNs for molecular geometry and transformers for SMILES sequences, improving molecular property prediction. UniGraph \cite{he2024unigraph} blends GNNs for structure with PLMs for text, enabling cross-domain transfer by integrating both data types.
\vspace{-5pt}
\subsection{Discussion} 

Different backbone models have their respective strengths, and it is unclear which is the best fit for GFMs. Currently, GNN-based models are strong contenders due to their expressiveness to effectively utilize structural information for graph tasks. As datasets continue to grow, transformer-based models are poised to take the lead due to their superior ability to capture the global graph structure compared to GNN-based models. Recently, LLM-based models, particularly those with instruction tuning, have shown remarkable performance in tasks like node classification and link prediction, demonstrating impressive flexibility \cite{LLMbenchmark}. As a result, LLM-based models are likely to play a larger role in future GFMs. Besides, hybrid models combine the strengths of different backbones, offering new possibilities for task and domain transfer. 

\section{Knowledge Acquisition}
\label{sec:4}
\vspace{-10pt}

Knowledge acquisition enables GFMs to extract meaningful patterns from graph data. with pre-training being the key approach. This process plays a crucial role in enhancing the transferability of GFMs, and the focus during pre-training differs depending on the specific type of transferability. For cross-task transferability, pre-training on data from various tasks helps the GFM better understand task-specific patterns, allowing it to learn the relevant task-related knowledge. However, since different tasks focus on distinct graph aspects, it is important to reduce the task gap during pre-training to ensure the model learns efficiently and effectively. For both intra-domain and cross-domain transferability, the datasets used for pre-training often exhibit structural and feature variances. As illustrated in Figure 1, general-purpose GFMs need to accommodate datasets from various graph sources in order to learn more transferable knowledge. Even within a single domain, the graph structures and features across different datasets can differ significantly. Therefore, mitigating the \emph{graph disparity} during pre-training is crucial for effectively learning transferable knowledge and preventing negative transfer \cite{GCOPE}.
In this section, we first introduce techniques to mitigate the \emph{task gap} and \emph{graph disparity} in pre-training. Following this, we provide a detailed overview of the pre-training methods commonly used in GFMs.

\vspace{-5pt}
\subsection{Mitigating Task Gap }
\label{sec:4.2}
Learning transferable task-specific knowledge during pre-training is key for efficient cross-task transferability. However, as different tasks focus on distinct graph aspects \cite{liu2023one}, mitigating the task gap during pre-training is essential. Subgraph sampling and multi-task learning are two effective strategies for acquiring relevant knowledge across tasks.

\vspace{-5pt}
\subsubsection{Subgraph sampling.}
\label{sec:4.2.1}
Many node and edge-level tasks can be generalized into operations at the graph level. For instance, tasks like “delete a subgraph” can essentially be viewed as “delete nodes and edges” \cite{sun2023all}. Therefore, subgraph sampling is a way to unify different graph tasks. 
In current GFMs, ProG \cite{sun2023all} converts node and edge tasks into graph-level tasks via subgraph sampling, while GCC \cite{qiu2020gcc} uses subgraph instance discrimination to capture universal patterns for transfer across node and graph classification tasks. Additionally, Prodigy \cite{huang2024prodigy} builds subgraphs to capture local context, enhancing input representations with structural information and supporting in-context learning. These examples demonstrate that subgraph sampling not only preserves key structural details but also improves the adaptability of models across various tasks.

\vspace{-3pt}
\subsubsection{Multi-task learning.}
\label{sec:4.2.2}

Multi-task learning uses multiple graph tasks during pre-training. This allows models to capture diverse patterns, such as node connectivity and global graph patterns. 
In current GFMs, Triplet-GMPNN \cite{ibarz2022generalist} is a generalized neural network for diverse algorithm tasks like sorting, searching, and geometric computations. It uses multi-task pre-training to improve input representation, training strategies, and architecture, enabling knowledge transfer across tasks. MiniMol \cite{klaser2024minimol} uses multi-task pre-training across molecular tasks, excelling in over 3,300 molecular benchmarks and demonstrating high transferability in molecular tasks. additionally, JMP \cite{jmp} advances the prediction of atomic properties by leveraging approximately 120 million atomic structures, treating each chemical dataset as an independent pre-training task to learn generalized atomic interaction representations that are highly transferable to various downstream tasks.

\subsection{Mitigating Graph Disparity}
\label{sec:4.1}

Learning diverse graph patterns during pre-training is essential for improving the transferability of both domain-specific and general-purpose GFMs. However, disparities in graph topologies and features across datasets pose challenges, as naive dataset aggregation may lead to negative transfer \cite{GCOPE}. Therefore, mitigating  graph disparity during pre-training is crucial. This is typically achieved by unifying the graph space, which involves aligning both features and topologies across datasets while strengthening their connections. We outline three commonly used methods below.

\vspace{-3pt}
\subsubsection{Aligning graph structures and features with SVD.}
\label{sec:4.1.1}
Singular Value Decomposition (SVD) is a powerful method in GFMs for aligning feature and topology dimensions across diverse datasets. Its strength lies in simplifying complex graphs while retaining essential structural and semantic patterns. In current GFMs, OpenGraph \cite{xia2024opengraph} and AnyGraph \cite{xia2024anygraph} both harness SVD to break down normalized adjacency matrices into a shared lower-dimensional space, ensuring topological consistency across datasets. Besides, ProG \cite{sun2023all} and GCOPE \cite{GCOPE} apply SVD to shrink initial feature dimensions to a uniform size, creating a unified feature space from varied sources. 

\vspace{-3pt}
\subsubsection{Using PLM for semantic alignment.}
\label{sec:4.1.2}
While SVD is good at aligning topological and feature dimensions across datasets, it has limitations in capturing the rich semantic relationships inherent in text attributes, especially when dealing with complex contexts or nuanced information. This gap has led to the increasing use of pre-trained language models (PLMs), such as DeBERTa \cite{he2020deberta}, XLNet \cite{yang2019xlnet}, and etc, to transform raw textual descriptions into embeddings, seamlessly mapping them into a unified vector space \cite{tape, giant, duan2023simteg, zhao2023gimlet}. In current GFMs, ZeroG \cite{li2024zerog} aims at improving cross-domain zero-shot transferability in graph learning. It uses SentenceBERT \cite{reimers2019sentence} to encode both node attributes and class descriptions into a shared semantic space. The shared semantic space allows ZeroG to seamlessly map new graph data to previously learned representations, enhancing its ability to generalize across diverse datasets without extensive retraining. 
Similarly, models like GraphAlign \cite{hou2024graphalign}, BooG \cite{boog}, and OFA \cite{liu2023one} utilize PLMs to embed textual node features into continuous vectors, promoting feature space consistency across varied datasets during pre-training.

\vspace{-5pt}
\subsubsection{Prompt as the bridge.}
\label{sec:4.1.3}

Graph prompt is a powerful tool for handling cross-domain graph data \cite{GCOPE, MDGPT, lachi2024graphfm, shirkavand2023deepgpt, gong2024selfproselfpro}, enhancing graph models to distinguish between structures and create a unified multi-domain feature space. In general, graph prompts are designed as ``virtual nodes'' or ``learnable tokens'' that connect with other nodes to enable dynamic information exchange \cite{sun2023graph}. For instance, MDGPT \cite{MDGPT} and ZeroG \cite{li2024zerog} use learnable tokens tied to specific source domains. These tokens serve as bridges between distinct semantic spaces, aligning node features from different domains. GraphFM \cite{lachi2024graphfm} employs a perceiver-based encoder with latent tokens, transforming domain-specific features into shared virtual nodes, creating a unified graph space. GCOPE \cite{GCOPE} introduces coordinators, which are virtual nodes that serve as dynamic bridges between datasets, harmonizing their features while maintaining individual properties. These coordinators foster collaboration both within and across graphs, promoting a unified yet flexible space.

\subsection{Pre-training Methods in GFMs}
\label{sec:4.3}

\textit{Section \ref{sec:4.1} and \ref{sec:4.2}} discuss how to organize data and formats to mitigate domain and task gaps during pre-training. In this section, we will focus on the specific pre-training methods. Graph pre-training has garnered substantial interest, with several surveys \cite{zhao2024survey, liu2022graph, wu2021self, xie2022self} already offering detailed reviews of related work. Here, we only focus on methods relevant to current GFMs, presenting key approaches to inspire future GFM design.

\subsubsection{Supervised learning.}
\label{sec:4.3.1}
Supervised learning is an effective pre-training strategy for GFMs when labeled data is available, as it directly uses ground-truth labels to learn precise, task-specific patterns, leading to better accuracy and generalization. Many GFMs adopt this approach. For example, GraphFM \cite{lachi2024graphfm} uses supervised learning to boost the model adaptability, enabling it to generalize effectively across a variety of graph datasets and tasks. Prodigy \cite{huang2024prodigy} employs supervised pre-training to directly resemble the format of downstream tasks, allowing it to perform well on node and edge classification tasks. For molecular graphs, Minimol \cite{klaser2024minimol} and JMP \cite{jmp} use rich labeled data and a multi-task framework to learn diverse structural patterns, enabling more comprehensive representations for both graph- and node-level tasks.

\subsubsection{Self-supervised learning.}
\label{sec:4.3.2}
The key limitation of supervised learning is its reliance on large amounts of labeled data \cite{liu2022graph}. Self-supervised learning (SSL) effectively overcomes this issue, making it especially valuable for GFMs, which often pre-train on diverse, unlabeled graph datasets \cite{wu2021self}. 
In general, SSL methods can be categorized into contrast-based and generation-based methods.
\noindent \emph{\underline{Contrast-based methods.}}
Contrast-based methods are grounded in maximizing mutual information by contrasting positive and negative samples \cite{hjelm2019}. Some GFMs directly adopt classic contrast-based methods for pre-training. For example, ProG \cite{sun2023all} employs GraphCL \cite{you2020graphcl} and DGI \cite{dgi}, while GCOPE \cite{GCOPE} utilizes GraphCL \cite{you2020graphcl} to enhance its graph representations. In contrast, many GFMs develop their own contrastive techniques. 
BooG \cite{boog} uses an anchor node as a positive sample before aggregation and treats other nodes in the batch as negatives, helping the model learn sharper, more discriminative representations by emphasizing similarities among related nodes and contrasts with unrelated ones. GraphGPT \cite{tang2024graphgpt} applies contrastive learning in a unique way by aligning text and graph structures, helping the model identify and reinforce correct graph patterns based on textual context. 
For molecular representation, COATI \cite{kaufman2024coati} employs contrastive learning to align 2D textual and 3D point representations of molecules, helping the model capture essential features of molecular structures in both forms. Models like AnyGraph \cite{xia2024anygraph}, MDGPT \cite{MDGPT}, ULTRA \cite{galkin2023towards}, and ULTRAQUERY \cite{ULTRAQUERY} employ link prediction for pre-training, reinforcing their ability to capture robust representations by maximizing the scores for actual links and minimizing them for unrelated node pairs.

\noindent \emph{\underline{Generation-based methods.}}
Generation-based methods focus on reconstructing the input data, using it as a form of self-supervision to capture structures and attribute semantics of graphs \cite{hou2022graphmae}. 
Though fewer GFMs adopt generation-based methods compared to contrast-based ones, some have found success. Opengraph \cite{xia2024opengraph} uses masked autoencoding, where parts of the graph are hidden, and the model learns to predict the missing pieces, capturing intrinsic structural patterns for better downstream performance. GCOPE \cite{GCOPE} utilizes a reconstruction module as a regularizer to align graph features across datasets, improving cross-domain transfer. Similarly, Unigraph \cite{he2024unigraph} applies Masked Graph Modeling to predict masked node features, enabling the model to generalize well across diverse graph structures and tasks.

\vspace{-3pt}
\subsection{Discussion}
\label{sec:4.4}
In current GFMs, a majority of models leverage multiple datasets during pre-training to facilitate intra-domain and cross-domain transfer. However, some GFMs \cite{boog, sun2023all} still pre-train on a single dataset. They employ alternative methods to bridge domain gaps effectively, which we will discuss in the next section. Besides, there is a lack of attention to intra-domain transferability, particularly regarding challenges such as size variance and data distribution shifts within the same domain. For specific pre-training methods, supervised and contrast-based learning are currently the most widely used. However, with the increasing scale of data, the high cost of data labeling is becoming a limiting factor. We expect more GFMs will shift towards contrastive or generative learning techniques to fully leverage data and acquire richer knowledge without relying heavily on labeled datasets.

\section{Knowledge Transfer}
\label{sec:5}
\vspace{-10pt}

Knowledge Transfer determines how well a GFM can apply its learned knowledge to downstream tasks or domains. Since different tasks focus on different aspects of graphs, and data distributions can vary significantly across domains, knowledge transfer is primarily concerned with overcoming these challenges to enable the effective and rapid application of learned knowledge. For both intra-domain and cross-domain transfer, the main challenge lies in data variances, where the structure and features of the datasets GFMs need to adapt to may differ greatly from those seen during pre-training. In such cases, graph prompts can help mitigate the differences between the learned data and downstream data. For cross-task transfer, the tasks to which GFMs must adapt might be entirely different from those encountered during pre-training. In this case, graph prompts combined with fine-tuning are effective methods for addressing the challenge. Therefore, in this section, we categorize knowledge transfer into prompt-based and finetune-based approaches.

\vspace{-6pt}
\subsection{Prompt-based Strategies}
\label{sec:5.1}

A key method for enhancing GFM knowledge transfer across domains and tasks is the use of graph prompts \cite{sun2311graph}. Graph prompts serve two main functions: 1) domain adaptation and 2) task adaptation. For domain adaptation, graph prompts are used to modify or adjust downstream data so that it aligns more closely with learned representations of the pre-trained model, thereby improving both intra-domain and cross-domain transferability. For task adaptation, graph prompts facilitate the transition between tasks by guiding the model focus, which enhances cross-task transferability and allows the model to seamlessly switch between various graph tasks.

\vspace{-5pt}
\subsubsection{Graph prompt for domain adaptation.}
\label{sec:5.1.1}
Some graph prompts are designed to adjust downstream graph data, making it more compatible with pre-trained models. 
For instance, ProG \cite{sun2023all} provides a comprehensive strategy by designing prompt tokens, structures, and insertion patterns to adjust downstream data for better model compatibility, enabling effective knowledge transfer between diverse domains such as citation networks and e-commerce. BooG \cite{boog} takes a different approach by integrating virtual super nodes, which act as information hubs to unify structural representations, promoting robust cross-domain knowledge transfer between citation networks and biology networks. Besides, MDGPT \cite{MDGPT} employs mixing prompt to incorporate domain-specific knowledge, creating a nuanced blend that aligns downstream data with the learned representations from the pre-trained model.

\vspace{-5pt}
\subsubsection{Graph prompt for task adaptation.}
\label{sec:5.1.2}

Current GFMs have designed a variety of prompts to enable effective cross-task adaptation. 
Models like OpenGraph \cite{xia2024opengraph} and AnyGraph \cite{xia2024anygraph} transform nodes into node pairs, where one prompt node represents candidate label classes and the other represents the real node entities. This approach effectively redefines node classification tasks as link prediction tasks, enabling pre-trained GNNs to adapt to new tasks with minimal fine-tuning. In addiyion, Prodigy \cite{huang2024prodigy} constructs a unified graph structure that integrates task-specific examples and queries as nodes, with connections based on their relationships and labels. This prompt graph structure allows the model to use contextual information from examples to accurately infer labels for the queries, enabling effective task adaptation without retraining. OFA \cite{liu2023one} facilitates cross-task transfer by appending task-specific prompt graphs, known as nodes-of-interest subgraphs, to the main graph. These prompt graphs and prompt class nodes connected by specific edge types that encode task semantics. This structure allows the model to synthesize information from support examples and task descriptions, thereby enhancing its ability to make accurate predictions across different tasks.

\subsection{Finetuning-based Strategies}
\label{sec:5.2}

Fine-tuning typically involves modifying the parameters of the pre-trained model itself. Vanilla fine-tuning is rarely used in current GFMs, as it can often lead to overfitting and unstable performance, especially when limited data is available \cite{yu2024few}. Instead, more efficient fine-tuning techniques are preferred. These methods require fewer data or fewer parameter updates, enabling few-shot or even zero-shot learning. In this section, we describe three common approaches: Meta-learning, Parameter-efficient fine-tuning (PEFT), and Probing.

\vspace{-5pt}
\subsubsection{Meta learning.}
\label{sec:5.2.1}
Mainstream meta-learning methods, like Model Agnostic Meta-Learning (MAML) \cite{finn2017model} and Prototypical Networks (Protonets) \cite{snell2017prototypical}, focus on extracting prior knowledge from a set of “meta-training” tasks that resemble the downstream “meta-testing” tasks. The goal of these approaches is to acquire initialization parameters or adaptable strategies that allow the model to be fine-tuned efficiently for new tasks, encapsulating the principle of “learning how to learn”. In current GFMs, ProG \cite{sun2023all} focuses on tailoring a globally shared model to better suit individual tasks, thus preserving the unique characteristics of each task while utilizing shared knowledge. It divides tasks at the node, edge, and graph levels for meta-training and meta-testing, and refines graph prompts iteratively across different tasks, enabling efficient knowledge transfer.

\vspace{-5pt}
\subsubsection{Parameter-efficient fine-tuning.}
\label{sec:5.2.2}
PEFT focuses on updating only a small subset of parameters of the model, while keeping the majority of the pre-trained model weights frozen. This approach reduces computational costs and minimizes the risk of overfitting when data is scarce \cite{zhao2024survey}. In NLP, popular PEFT methods include adapter-tuning \cite{houlsby2019parameter}, BitFit \cite{zaken2021bitfit}, Low-Rank Adaptation (LoRA) \cite{hu2021lora}, and etc. 
Inspired by these methods, graph PEFT has recently gained attention. In current GFMs, UniGraph \cite{he2024unigraph} uses LoRA to enable efficient knowledge transfer with minimal parameter updates, adapting to various graph tasks while freezing most model parameters. Uniquely, MiniMol \cite{klaser2024minimol} generates molecular fingerprints from the final layer of the pre-trained model, using these embeddings to train a smaller MLP for specific downstream tasks, offering a lightweight alternative to full model fine-tuning.
\vspace{-5pt}
\subsubsection{Probing.}
\label{sec:5.2.3}
Probing typically involves attaching a lightweight “probe” network, which can take various forms, such as an additional linear layer (referred to as linear probing), a multi-layer perceptron (MLP), or even a non-parameterized mapping function \cite{zhao2024survey}. During the fine-tuning stage, only the probe is trained, allowing for efficient and lightweight knowledge transfer. 
In the context of GFMs, probing is frequently employed. For instance, GraphFM \cite{lachi2024graphfm} and GCC \cite{qiu2020gcc} both apply linear probing methods, with GraphFM utilizing a linear classifier and GCC adopting logistic regression or SVM. These methods are simple yet effective, focusing on rapid transfer with minimal data. Methods like BooG \cite{boog} and ProG \cite{sun2023all} extend the linear probing by adding MLP layers for few-shot learning, offering more flexibility to capture complex task-specific knowledge. Besides, COATI \cite{kaufman2024coati} trains differentiable regressors to further fine-tune its pre-trained models. 
However, the simplicity of the probe architecture may restrict the performance on more complex downstream tasks where a more nuanced understanding of graph structures or relationships is required.

\section{Discussions and Future Directions}
\label{sec:7}
\vspace{-10pt}

\textbf{Domain-Specific vs. General-Purpose GFMs.} 
At the current stage, domain-specific GFMs are more feasible as they focus on well-defined tasks within a specific domain, leveraging structured knowledge and curated datasets. This specialization allows for high efficiency and strong intra-domain transferability but limits adaptability across domains. On the other hand, general-purpose GFMs promise greater versatility by aiming to generalize across multiple domains and tasks. However, achieving strong cross-domain transferability remains challenging due to significant differences in graph structures, feature distributions, and data availability. A potential direction worth investigating is whether domain-specific GFMs can contribute to the development of more effective general-purpose GFMs. Instead of training a general-purpose model from scratch, multiple domain-specific GFMs could be integrated to form a modular architecture, where different components specialize in distinct domains but share a common knowledge representation. Furthermore, if general-purpose GFMs become feasible, efficient adaptation mechanisms will be essential to maximize their utility for domain-specific applications. While a general-purpose model can serve as a strong starting point, it must still be tailored to specific domains to ensure optimal performance.

\noindent\textbf{The Role of LLMs in GFMs.} 
Recent studies have shown that LLMs can effectively handle graph-related tasks, such as node classification, link prediction, and graph reasoning, by leveraging their strong pattern recognition and representation learning abilities~\cite{LLMbenchmark, chen2024llaga, InstructGLM, tang2024graphgpt, zhu2025llm, wu2025comprehensive, sun2025graphicl, wang2023can}. However, they still face challenges in capturing complex relational structures and scalability to large graphs. Integrating LLMs with graph models could combine their reasoning and generalization strengths with the structural awareness of traditional graph learning methods. Developing hybrid models that unify LLM-driven semantic understanding with graph-based relational learning presents a promising research direction for improving GFM adaptability and cross-domain transferability.

\noindent\textbf{Standardized Benchmarking for Transferability.} The lack of standardized benchmarking for GFM transferability hinders progress. Comprehensive benchmarks are needed to assess GFM performance across domains and tasks, helping identify key factors that influence transferability. Establishing such benchmarks will drive GFM advancements and pinpoint areas for improvement.

\section{Conclusion}
\vspace{-10pt}

This survey provides a comprehensive review of GFMs from a transferability perspective, introducing a systematic taxonomy that categorizes existing GFMs based on Intra-Domain, Cross-Domain, and Cross-Task Transferability, as well as their application scope (Domain-Specific vs. General-Purpose). 
Following this taxonomy, we outline a general framework consisting of Backbone Models, Knowledge Acquisition, and Knowledge Transfer, highlighting key factors that influence GFM transferability. By examining current progress and challenges, we provide a structured perspective on the development of GFMs and their applications across various domains. We hope this survey serves as a foundation for future research, guiding the advancement of more effective and transferable GFMs for diverse real-world applications.

\newpage
\bibliographystyle{ACM-Reference-Format}
\bibliography{KDD-survey/GFM}



\end{document}